\documentclass[letterpaper, paper,11pt]{AAS}	

\usepackage{bm}
\usepackage{amsmath}
\usepackage{amssymb}

\usepackage[colorlinks=true, pdfstartview=FitV, linkcolor=black, citecolor= black, urlcolor= black]{hyperref}
\usepackage{overcite}
\usepackage{footnpag}			      	

\usepackage{graphicx}
\usepackage{siunitx}
\usepackage{longtable,tabularx}
\usepackage{subcaption}
\usepackage{soul}
\usepackage{placeins}

\PaperNumber{24-445}

\begin{document}

\title{Fine-Tuned Language Models as Space Systems Controllers}

\author{Enrico M. Zucchelli\thanks{Postdoctoral Associate, Aeronautics \& Astronautics, Massachusetts Institute of Technology, MA 02139.}, Di Wu\footnotemark[1], Julia Briden\thanks{PhD Candidate, Aeronautics \& Astronautics, Massachusetts Institute of Technology, MA 02139.}, Christian Hofmann\footnotemark[1], Victor Rodriguez-Fernandez\thanks{Associate Professor, Department of Computer Systems Engineering, Universidad Politécnica de Madrid, 28040 Madrid, Spain}, Richard Linares\thanks{Associate Professor, Aeronautics \& Astronautics, Massachusetts Institute of Technology, MA 02139.}, 
}

\maketitle{}

\begin{abstract}

Large language models~(LLMs), or foundation models~(FMs), are pretrained transformers that coherently complete sentences auto-regressively. In this paper, we show that LLMs can control simplified space systems after some additional training, called fine-tuning. We look at relatively small language models, ranging between 7 and 13 billion parameters. We focus on four problems: a three-dimensional spring toy problem, low-thrust orbit transfer, low-thrust cislunar control, and powered descent guidance. The fine-tuned LLMs are capable of controlling systems by generating sufficiently accurate outputs that are multi-dimensional vectors with up to 10 significant digits. We show that for several problems the amount of data required to perform fine-tuning is smaller than what is generally required of traditional deep neural networks (DNNs), and that fine-tuned LLMs are good at generalizing outside of the training dataset. Further, the same LLM can be fine-tuned with data from different problems, with only minor performance degradation with respect to LLMs trained for a single application. This work is intended as a first step towards the development of a general space systems controller.
\end{abstract}

\section{Introduction}

Large language models (LLMs), also known as foundation models (FMs), are statistical models that interact through language. The recent years have seen a proliferation of these models, with ever-increasing size, variety, and capabilities. Several existing commercial companies have developed their own LLMs, and many more ventures have been born from this technology alone. Further, some models can process multiple modes of inputs other than text, including sound and video\cite{team2023gemini}. Through language, LLMs, have displayed the ability to control agents and interact with the world~\cite{xu2023drivegpt4,rodriguez2024language,sinha2024real}. The benefits of using language models over more traditional forms of artificial intelligence~(AI) include explainability, better handling of rare events, good zero-shot and few-shot performance, as well as improved pattern recognition~\cite{mirchandani2023large}.
These features emerge from extensive and comprehensive pretraining, during which the FM gains knowledge of the world under the form of text and other modes; this in turn allows the FM to recognize abstract or remote similarities between a new input and the data it has been pretrained on.
The goal of this paper is to build a first step towards a general space controller that is capable of completely autonomous control. The long-term end of this line of research would be that it would only require the desired high-level goal in text form as input.

LLMs exploit the attention mechanism of the transformer architecture~\cite{vaswani2017attention}, which is the key feature at the core of the recent revolution in artificial intelligence~(AI) models.
Before public release, FMs are trained with terabytes of text data gathered from the internet. The data is used to adjust the network's internal parameters, the number of which can range between one billion and more than one trillion, depending on the model. At every step, the LLM is given an input consisting of a sequence of tokens (a token is either a word, a subword, or a character), and it is trained to estimate a probability distribution for the tokenfollowing the input sequence of tokens. When used for inference, a token is sampled from the predicted output distribution, and then added to the input sequence in an auto-regressive manner. The LLM then completes sentences and generates new text, possibly outside of its training data. In this sense, LLMs belong to the class of generative AI. 
Additional fine-tuning can be performed to obtain task-specific models. The fine-tuned model can now be adapted to various goals, such as interacting in the form of a chatbot.

Recently, GPT4, one of the most advanced networks available to date, has shown the ability to drive cars~\cite{xu2023drivegpt4} and to operate spacecraft~\cite{rodriguez2024language}, with little to no additional data or training. LLMs have three advantages over other AI approaches: 1) they can generally learn with less data, leveraging the knowledge of the world already compressed within their parameters (zero-shot and few-shot performance), 2) they can handle rare events or situations that may have little in common with the cases they have been exposed to during training through ``common sense", which too derives from general experience obtained during pretraining, and 3) since they use language, they are capable of providing explanations for their actions (albeit hallucinations are sometimes possible). Finally, 4) researchers have shown that large pretrained models excel at pattern recognition~\cite{mirchandani2023large}, even when such pattern has likely nothing in common with pretraining data. This ability can implicitly emerge because of the complexity of the networks and the large amount of data provided. In their work, Mirchandani et al. show these capabilities while in-context, \textit{e.g.}, without performing any fine tuning but just by providing input sequences; in this work, instead we leverage the pattern recognition capabilities of LLM during fine-tuning, \textit{e.g.}, by changing the parameters of the models.

This work explores how large pretrained transformers can solve space-related problems, such as optimal orbit transfer, maneuvering in the cislunar region, and 3 degrees-of-freedom~(DoF) landing problems. We focus on relatively small and open-source architectures, such as Meta's Llama-2~\cite{touvron2023llama}, and aim to minimize the data required to control space systems successfully. This research aims to be an effort in the direction of obtaining an agent capable of controlling a spacecraft in a general setting and under various unexpected situations.

\section{Large Language Models}
An LLM is a computational model that generates general-purpose language. In an LLM, text is first transformed into a list of indices, called tokens~\cite{grefenstette1999tokenization}: the text is divided into small segments of string, each corresponding to a word or a subword, and each segment is given an index number, called token. This process is called tokenization, and is performed by a deterministic function that is not trained on data. For some models, a variety of 100,000 tokens or more are available. The model can take as input a maximum number of tokens (the maximum number of input tokens processed at once is called context length), which for GPT-3\cite{radford2018improving} is 2,048 but can be as large as several millions, such as the case of Google's Gemini 1.5\cite{reid2024gemini}. Each token is then transformed into a multi-dimensional vector (of length 12,288 for GPT-3\cite{radford2018improving}), called the embedding~\cite{patil2023survey}. The function that transforms a token index into a multi-dimensional vector is trainable. Tokens with similar meanings are encoded to similar vectors, and the similarity is learned during pretraining, without any labeling required. As the token's position affects a word's meaning, a successive step, called positional encoding, is computed for every position and added to the embedding. After embedding, the sequence of integers is transformed into a sequence of high-dimensional differentiable vectors. Differentiability is a core requirement since all training happens through gradient descent~\cite{kingma2014adam}.

\subsection{Attention Mechanism}
Attention\cite{vaswani2017attention} is the key component of transformers and has heavily revolutionized the field of AI since 2017.
Attention is the process of generating multiple queries for every embedding in the input. Each query results from a multiplication between a query matrix and an embedding. A key matrix is associated with a query matrix, similarly generating a key for every embedding. For GPT-3\cite{radford2018improving}, query and key vectors all have a length of 12,288. Queries and keys can find semantic connections between words that are far apart from one another. One can then obtain a new layer of values:
\begin{equation}
    \text{Attention}\left(Q,K,V\right) = \text{Softmax}\left(\frac{QK^T}{\sqrt{d_k}}\right)V,
\end{equation}
where $Q$ is the matrix of all query vectors in the context (not to be confused with the query matrix itself), $K$ is the matrix of all keys, $V$ is the value matrix, which transforms each embedding to a normalized update embedding, and $d_k$ is the dimension of the key/query vector. The softmax operation is to be applied independently for every column of the resulting matrix. An attention layer for a specific query thus provides a value between 0 and 1, used to update the subsequent embedding. When performing many queries at once, multi-headed attention is required, in which the attention operation is performed multiple times in parallel, all with their own sets of key, query, and value matrices. In a large transformer such as GPT-3, 96 heads are used\cite{radford2018improving}. All the parameters of the query, key, and value matrix are tunable, and differ for different heads.

\subsection{LLM Architecture}
Architectures of LLMs can differ widely, but they generally include the same major blocks: 
tokenization, embedding, attention blocks, and feed forward nonlinear networks, as well as many normalization layers. In the nonlinear layers, the inputs to the layers are all multiplied by weight matrices and then passed through nonlinear activation functions. Depending on the architecture, some layers are repeated multiple times. The final output is a vector where to each token corresponds a number; after a softmax operation, one has the probability distribution for the next token.

\subsection{Quantization}
A consumer-grade graphic processing unit~(GPU) such as Nvidia's GeForce 4090 has a random-access memory~(RAM) of 24~GB. As such, it can fit at most 12 billion half-precision parameters. It has been shown that reducing the precision of the parameters comes with little cost~\cite{dettmers20218,frantar2022gptq}. For example, reducing the precision to 8~bit or even 4~bit is possible with a marginal reduction in accuracy; the lost accuracy is smaller for larger networks. This way, a compression of a factor of 4 can be obtained.

\section{Fine-Tuning}

A pretrained transformer can be fine-tuned to better adapt to the task it needs to perform. In this research, we are interested in control outputs, but other reasons for fine-tuning might be generating a chatbot such that it learns to answer questions or summarize text. Fine-tuning consists of changing the weights of the original pretrained transformer by exposing it to new, additional data, and performing stochastic gradient descent.

Pretraining and fine-tuning rely on back-propagation~\cite{lecun1988theoretical}, which has been the workhorse of supervised machine learning since its inception,  and stochastic gradient descent. The training methodology is equally valid for all AI statistical models, including feed-forward neural networks, convolutional neural networks~(CNNs), recurrent neural networks~(RNNs), and transformers. With stochastic gradient descent an NN is provided a set of input parameters for which the desired output is known. The difference between the NN's predicted output and the target output can be quantified with a loss function. The gradient for all the internal parameters is computed via chain rule. The approach is stochastic because, at each iteration, the gradient is computed for a randomly selected subset of the data, instead of all the data; computing the gradient for all data at once would be too computationally demanding. Further, stochasticity is generally beneficial in avoiding local minima.
The most commonly used algorithm is Adam\cite{kingma2014adam}, which is memory intensive as it needs to store the gradient and two additional vectors, each of the same size as the number of parameters of the LLM. Hence, we utilize low-rank fine-tuning to reduce the memory footprint. As an ancillary benefit, if one is training several fine-tuned models, the data to be stored is also reduced.

\subsection{Low-Rank Adaptation}
Low-Rank adaptation\cite{hu2021lora}~(LoRA) works by compressing the weight updates of a layer as a low-rank matrix:
\begin{align}
    W^+ &= W^- + \Delta W,\\
    \Delta W &= B\,A,
\end{align}
where $W$, $\Delta W\in\mathbb{R}^{n_d\times n_d}$, $B\in\mathbb{R}^{n_d\times n_r}$, and $A\in\mathbb{R}^{n_r\times n_d}$, with $n_d$ the dimension of the layer, and $n_r\ll n_d$ the rank of the update matrix. 
Low-Rank Adaptation can be performed for virtually every matrix appearing in the inner workings of the LLM, and 

\section{Implementation}
Training and inference for this work has been performed using Hugging Face~\cite{jain2022hugging}, an open-source platform for machine learning. 

The inputs and outputs of an LLM consist of language. One needs to automatically generate prompts that convey, at the very least, the available information of the state.
In all the evaluated problems, a prompt describing the system's current state is automatically generated. It is then provided to the transformer, which returns a sentence that includes the numeric output. Especially when using little data, the more informative the prompt, the better the LLM can leverage its pretraining.

The output of an FM is by definition a string. It is therefore paramount that one implements methods to extract the desired data automatically. With the fine-tuning used in this work, this does not need to be optimized much, since the reply follows a very specific pattern for every data point.

Below is an example of prompt used in this research:\\
\begin{center}
\fbox{\begin{minipage}{35em}
\textbf{User}: You are an AI navigation system for a spacecraft. Your task is to guide the spacecraft from its current position to a specified destination without any time constraints (but within a single revolution) while minimizing the integral of the thrust squared.\\
\\
Input:\\
- Current spacecraft state: 3D position [0.703740, -1.317800, -0.077790] and velocity [0.823310, 0.088951, 0.119410]\\
- Destination coordinates [2.000000, 0.000000, 0.000000] with velocity [0.000000, 0.707107, 0.000000]\\
\\
The output should be a series of 3D thrust vectors [Tx, Ty, Tz] to be applied at specific times to navigate the spacecraft from its current position to the destination while minimizing the integral of the thrust squared.\\
\\
Constraints:\\
- The spacecrafts motion is governed by Newtonian mechanics in a vacuum.\\
- The spacecraft has a limited amount of fuel, so the integral of the thrust squared must be minimized to conserve fuel.\\
\\
\\
\textbf{AI}: Thrust vector to be applied at this time: [0.035448, 0.018499, -0.112130].
\end{minipage}}
\end{center}

\begin{figure}[htp]
    \centering
    \includegraphics[trim={1.6cm 1cm 2cm 0.6cm},clip,width=0.9\linewidth]{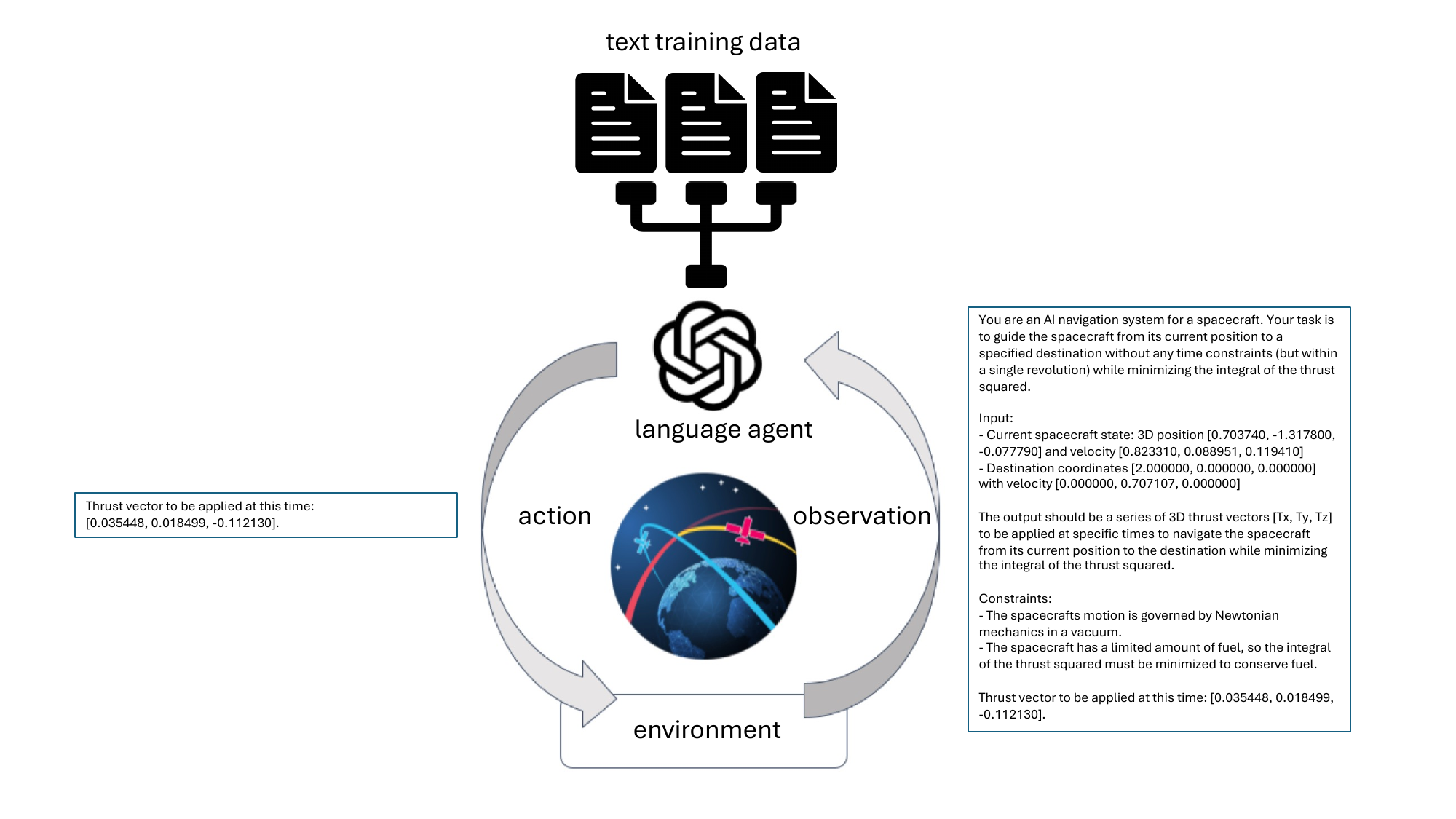}
    \caption{Illustration of the training and control process used in this work.}
    \label{fig:llmagent}
\end{figure}

A schema for the methodology used in this work is shown in Fig.~\ref{fig:llmagent}. Training data in the form of text is used to fine-tune the language agent. Then, text is automatically generated to translate the observations of the environment into text. The prompted language model provides text as output, which in turn is translated as inputs to the environment. Following interactions with the environment, new observations are generated. The text in the figure is the same as the text in the preceding textbox.

\section{Linear 3-Dimensional Spring}
For this toy problem, we train the LLM to bring a linear 3-dimensional mass-spring-damper system (with negative damping, causing instability in the problem) to the origin.
The linear system evolves according to the following ordinary differential equation:
\begin{equation}
    \dot{\bm{x}} = A\bm{x} + B\bm{u},
\end{equation}
where $\bm{x}$ is the stacked position and velocity of the spring, with
\begin{equation}
 \nonumber   A = \left[\begin{matrix}
        0 & I_{3\times3} \\
        -\frac{k}{m}I_{3\times3} & -\frac{c}{m}I_{3\times3}
    \end{matrix}\right],\qquad
    B = \left[\begin{matrix}
        0  \\
        \frac{1}{m}I_{3\times3}
    \end{matrix}\right],
\end{equation}
where $m$ is the mass of the object, $k$ is the stiffness, and $c$ is a damping coefficient.
The cost is
\begin{equation}
    J = \int_{0}^\infty \left(\bm{x}^TQ\bm{x} + \bm{u}^TR\bm{u}\right) dt,
\end{equation}
with $Q=I_{6\times 6}$ and $R = 0.1\,I_{3\times 3}$.

Optimal trajectories starting from randomly sampled initial conditions are generated using a linear quadratic regulator~(LQR). As there is no final time, the infinite-horizon solution can be employed:
\begin{subequations}
    \begin{equation}
        \bm{u} = - K\bm{x},
    \end{equation}
    \begin{equation}    
        K = R^{-1}B^T P,    
    \end{equation}
\end{subequations}
and $P$ is the solution to the algebraic Riccati equation:
\begin{equation}
    A^TP+PA-PBR^{-1}B^TP +Q=0.
\end{equation}
Tha dataset is discretized every 0.05~s. Note that the LQR solution is slightly suboptimal, since it is the optimal solution for the continuous-time problem, but it is applied to a discrete-time problem.

\subsection{Results}
In this setting, the LLM is interrogated with a frequency of 20~Hz. The initial conditions for the dataset trajectory are
drawn from a 6-dimensional normal distribution.
Table~\ref{tab:performance_comparisonlqr} summarizes single trajectory performance obtained as a ratio with respect to the LQR solution.
With data from only 3~trajectories, a fine-tuned Llama-2-13B is capable of stabilizing the system and bringing it to the origin, with a total cost that is 59\% larger than optimal. With data from just 30~trajectories the performance is as good as that of an LQR for initial conditions in distribution. Further, the right column of the table shows that the LLM can stabilize the system even when starting from points well outside of the training data, albeit sub-optimally. This is true for both the solutions used as dataset, as for the solutions used as comparison.

\begin{table}[h]
    \centering
    \begin{tabular}{|c|c|c|}
        \hline
        {Dataset size (number of trajectories)} & $J_{LLM}/J_{opt}$ & $J_{LLM}/J_{opt}$, 10$\sigma$ initial conditions\\
        \hline
        1  &  12.17 & 16.98 \\
        \hline
        3  &  1.59 & 5.98 \\ 
        \hline
        30  &  1.00 & 1.04 \\ 
        \hline
        300  &  1.00 & 1.05 \\ 
        \hline
        3000  &  1.00 & 1.02\\
        \hline
    \end{tabular}
    \caption{Result summary for the linear problem, using Llama-2-13B and differently sized datasets.}
    \label{tab:performance_comparisonlqr}
\end{table}

\section{Optimal Orbital Transfer}
\label{sec:OrbitalTransfer}
In this nonlinear problem the LLM is trained to learn the optimal control function in the surroundings of a reference optimal orbit transfer trajectory.
For this problem the only forces involved are the central gravity field and the thrust $\bm{u}(t)$:
\begin{equation}
    \Ddot{\bm{r}}(t) = - \frac{\mu}{r(t)^3}\bm{r}(t)+\bm{u}(t),
\end{equation}
where $\bm{r}$ is the position of the spacecraft and $\mu$ is the gravitational parameter of the central body.
The transfer involves simultaneous change of semi-major axis $a$, eccentricity $e$, and inclination $i$. Optimality is in the sense of minimum energy:
\begin{equation}
    J = \int_{t_0}^{t_f} \frac{1}{2} \bm{u}(t)^T R \bm{u}(t)\, dt,
\end{equation}
with $R=I_{3\times 3}$, and the initial and final boundary conditions are fixed values of the state. In this problem we utilize normalized distance units~(DU) and time units~(TU). The units are chosen such that an orbit at distance DU has period 2$\pi$ TU. 
The semi-major axis increases from 1.268~DU to 2~DU, the orbit is circularized from an initial eccentricity of 0.12, and the inclination is brought to 0 degrees from an initial value of 26.6 degrees.
The final time is free, but the transfer is forced to be completed within a single revolution.
The optimal trajectories used for training are generated by indirect optimal control, which transforms the original infinite-dimensional optimization problem into a two-point boundary value problem. Using primer vector theory~\cite{russell2007primer}:
\begin{subequations}
    \begin{equation}
    \dot{\bm{\lambda}}_r(t) = \mu\,\bm{\lambda}_v(t)^T \left(\frac{I_{3\times 3}}{\sqrt{r(t)^3}}-3 \frac{\bm{r}(t)\bm{r}\,(t)^T}{\sqrt{r(t)^5}}\right)
    \end{equation}
    \begin{equation}
    \dot{\bm{\lambda}}_v(t) = - \bm{\lambda}_r(t),
    \end{equation}
\end{subequations}
where $\bm{\lambda}_r$ and $\bm{\lambda}_v$ are the costate vectors for position and velocity, respectively, and
\begin{align}
    \bm{u}(t) = -\bm{\lambda}_v(t).
\end{align}
By finding $\bm{\lambda}_r(t_0)$ and $\bm{\lambda}_v(t_0)$ such that the constraints on the final state are met, one automatically obtains an extremal solution to the problem. In this problem, the values are sought by single shooting, using Levenberg-Marquardt and the trust region method~\cite{conn2000trust}.
An additional transversality condition is included for a time-independent, free final time problem:
\begin{equation}
    \frac{1}{2} \bm{u}(t_f)^T R\, \bm{u}(t_f) + \bm{\lambda}_r^T(t_f)\,\bm{v}(t_f)+\bm{\lambda}_v^T(t_f)\,\Ddot{\bm{r}}(t_f)  = 0,
\end{equation}
where $\bm{v}$ is the velocity of the spacecraft.
A dataset is generated by perturbing the initial conditions, both velocity and position, according to a 3-dimensional Gaussian distribution with standard deviation of 0.05 DU for the initial position and 0.05~DU/TU for the initial velocity. The trajectories that solve the problem with either 0 or 2 revolutions are pruned. Each trajectory is sampled 500 times during the transfer, and state-action pairs are saved. The training data consists of prompts where the user specifies the state of the spacecraft, and the LLM is trained to provide the corresponding thrust vector.

\subsection{Simulation Results}

\begin{figure}[htp]
    \centering
    \includegraphics[trim={0cm 6cm 0cm 7cm},clip,width=0.48\linewidth]{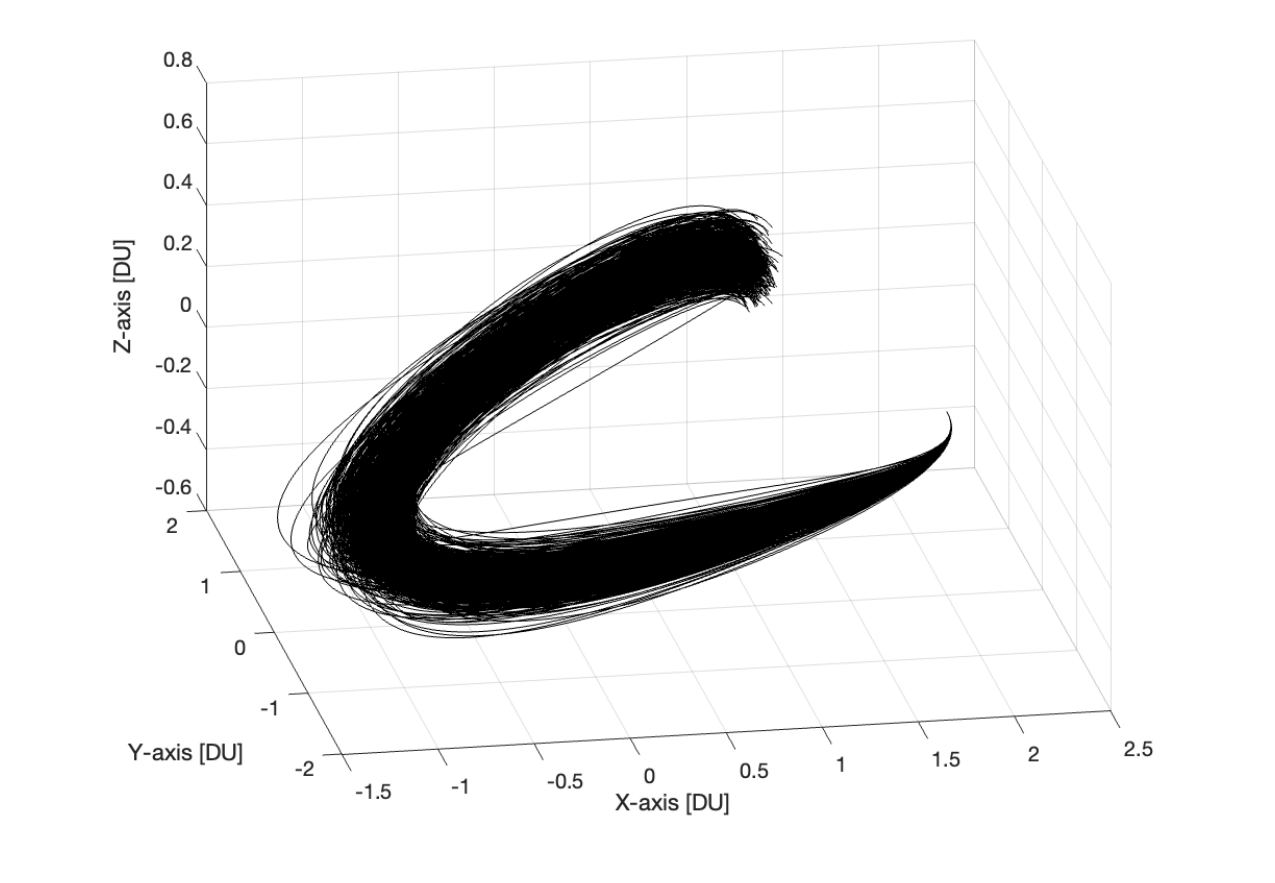}
    \includegraphics[trim={0cm 6cm 0cm 7cm},clip,width=0.48\linewidth]{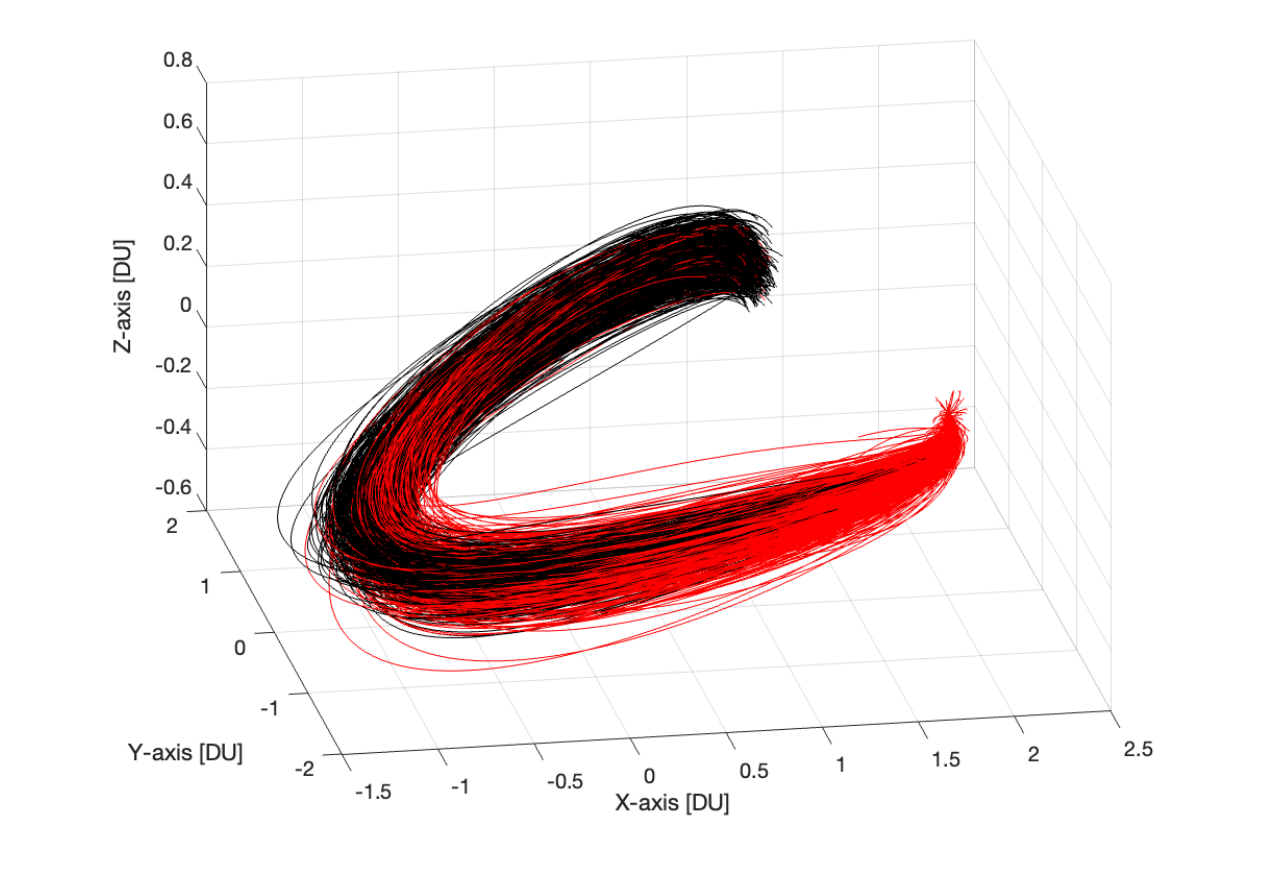}
    \caption{Dataset used for the orbital transfer problem: trajectories starting from perturbed initial conditions~(black, left), and all trajectories~(right), including trajectories starting along the mean trajectory~(red), then perturbed.}
    \label{fig:orbitraisedataset}
\end{figure}

Two different foundation models have been fine-tuned for this problem: Llama-2-7B and Llama-2-13B. Each has been trained with four different datasets. The full dataset comprises of 800 trajectories starting from the initial state, perturbed with a standard deviation of 0.05 for all states, and another 800 trajectories starting from random times along the central trajectories, and also perturbed with a standard deviation of 0.05. All trajectories in the dataset are obtained via the shooting method, with initial guess for the costates equal to zero. Because of the requirement of a good initial guess, the optimizer would converge to the optimal solution only 88.4 \% of the times. Convergence generally fails either because the final trajectory does not respect the constraint of only one revolution around the central body, or because of poor conditioning of the Jacobian matrix.
The dataset is illustrated in Fig.~\ref{fig:orbitraisedataset}. The red trajectories (in the right plot only) are the ones starting from along the reference trajectory. The addition of the red trajectories allows better behavior in the final portions of the trajectory, where the cone of data becomes very thin, and a small perturbation can take the spacecraft outside of the training data. Reduced datasets have been generated by selecting only a subset of the trajectories, chosen at random. The two LLMs have been fine-tuned using datasets of 25, 100, 400, and 1,600 trajectories. Each trajectory contains 500 time-steps, and every single time step is a separate data point for the fine-tuning. In every case, the total number of training steps has been kept constant, and equal to 800,000. Note that while the trajectories have been generated assuming continuous-time control, when tested, the control time is discretized. Hence, the agents are not expected to obtain the same performance as the dataset. For this problem, the exact prompt showed in Fig.~\ref{fig:llmagent} has been used. The agent receives prompts where the state contains up to 7 significant digits, and it is trained to provide outputs that also contain 7 significant digits.

\begin{figure}[htp]
    \centering
    \includegraphics[trim={0cm 6cm 0cm 7cm},clip,width=0.48\linewidth]{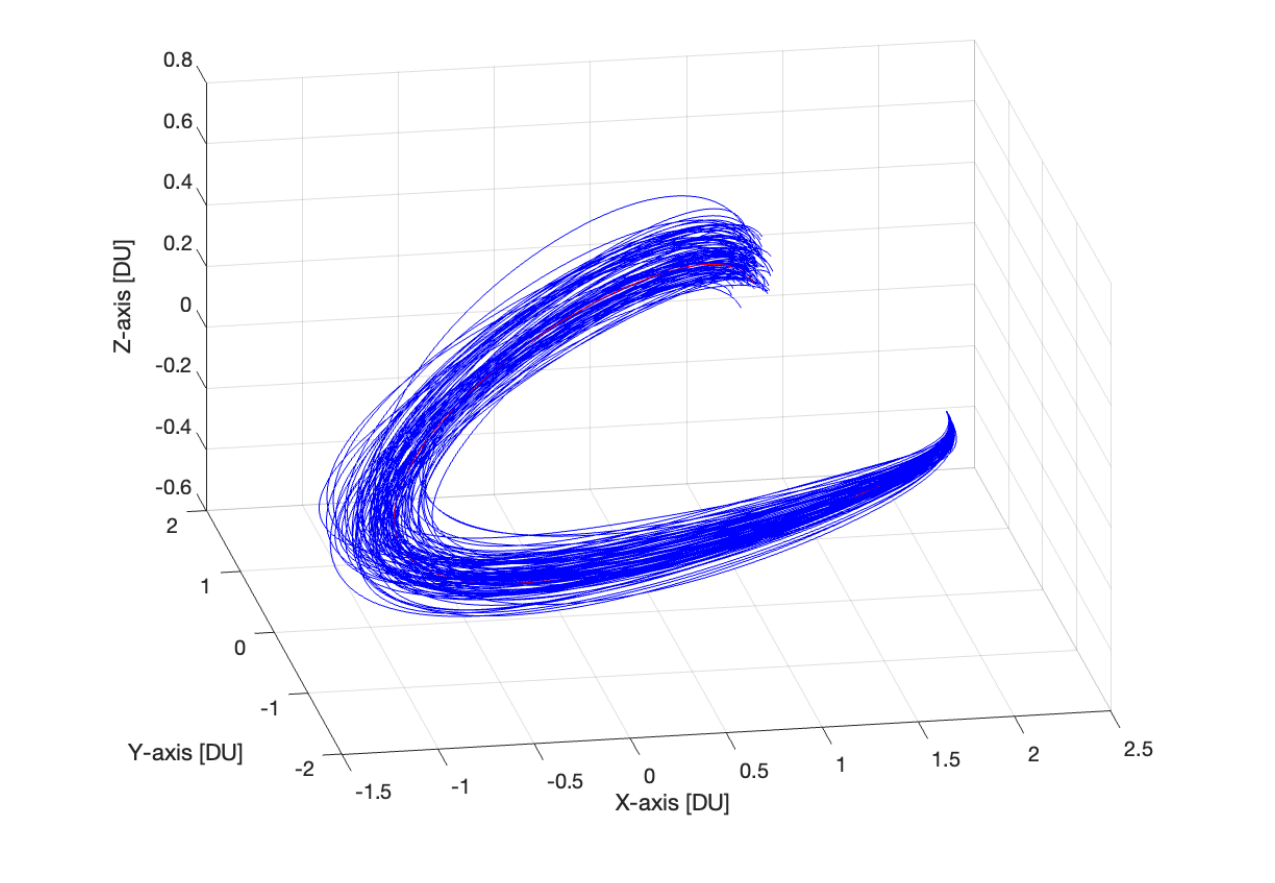}
    \includegraphics[trim={0cm 6cm 0cm 7cm},clip,width=0.48\linewidth]{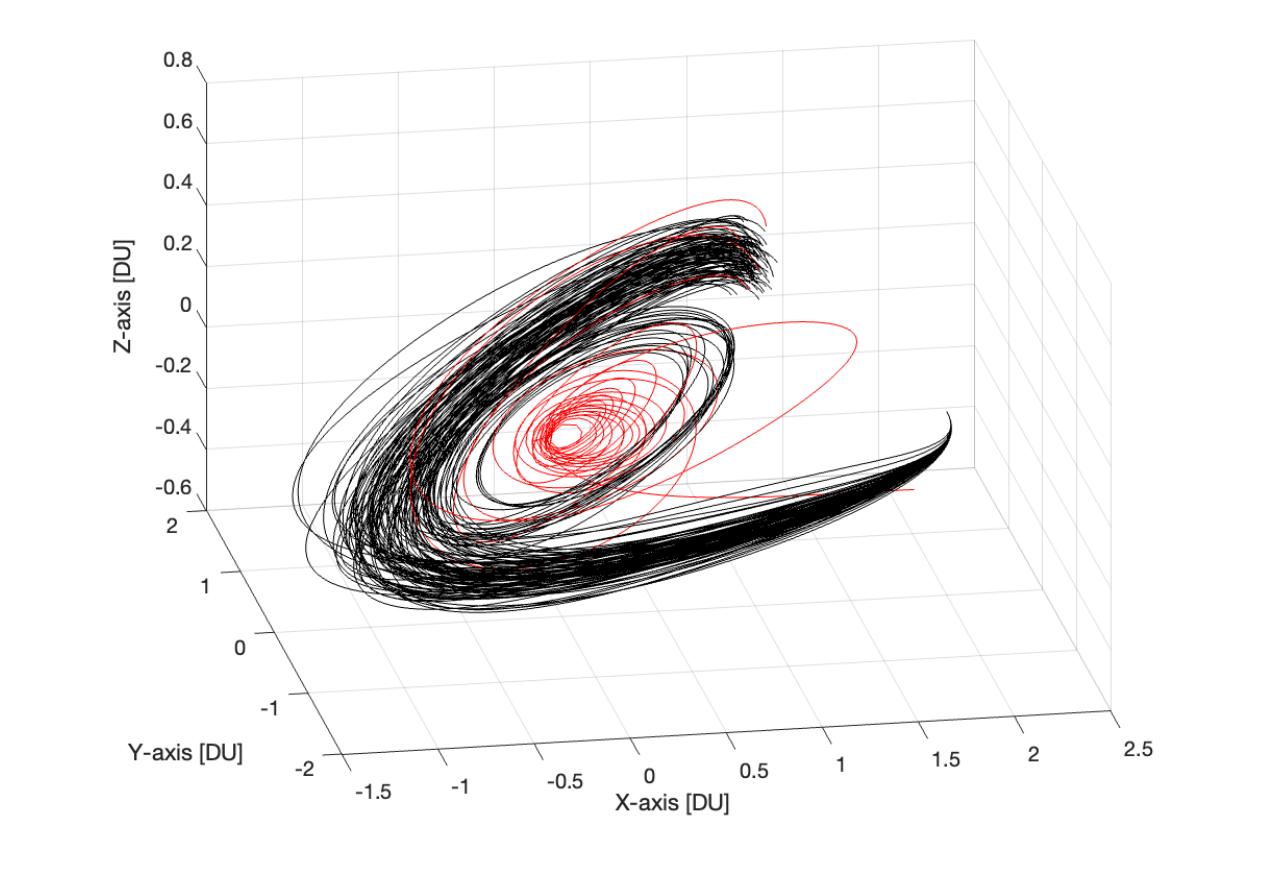}
    \caption{Trajectories guided by the Llama-2-7B-1600 (left), and trajectories computed with the optimizer (right), with four non-converging trajectories (red), and 13 trajectories that do not respect the single revolution constraint.}
    \label{fig:orbtransf_1600_7B_ICuB}
\end{figure}

All the LLMs have been tested over 100 different initial conditions, randomly picked from the same distribution of the dataset. The LLMs are called every 0.02~TU.
Figure~\ref{fig:orbtransf_1600_7B_ICuB} shows a comparison between the trajectories obtained with the LLM as a guidance logic on the left, and the trajectories obtained by optimal control on the right. While the LLM always achieves the desired target, the optimal control method generates 4 non-converging trajectories and 13 trajectories that take 2 or more revolutions, and thus a success rate of only 83\%.

\begin{figure}[htp]
    \centering
    \includegraphics[trim={4cm 8cm 4cm 9cm},clip,width=0.48\linewidth]{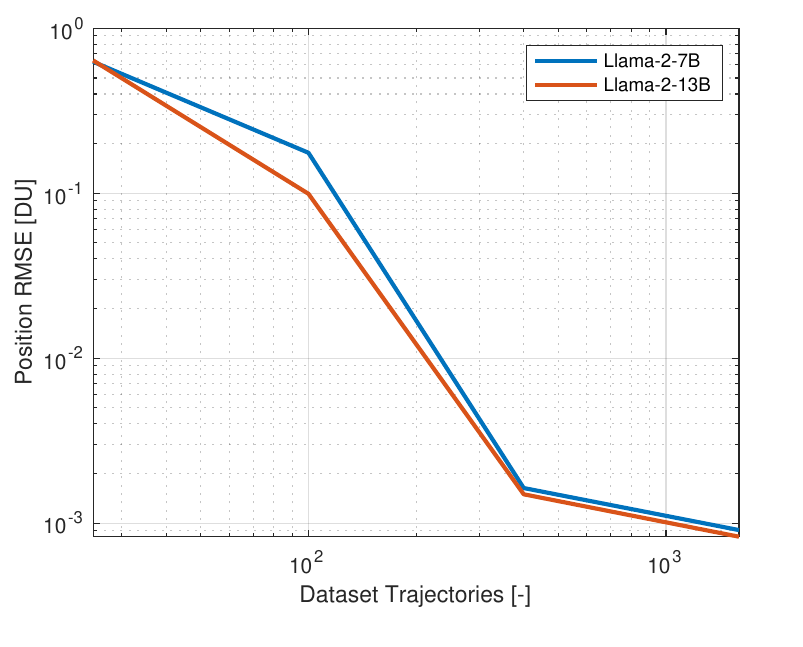}
    \includegraphics[trim={4cm 8cm 4cm 9cm},clip,width=0.48\linewidth]{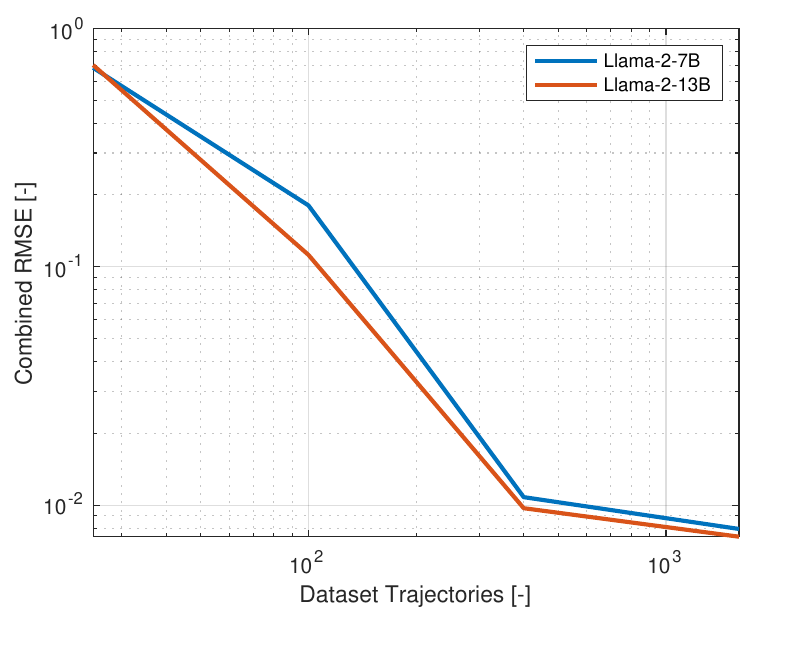}
    \caption{Position (left) and combined position and velocity (right) accuracy as a function of dataset size.}
    \label{fig:orbitTransferPosAllLLLMs}
\end{figure}

Figure~\ref{fig:orbitTransferPosAllLLLMs} shows the accuracy of the two different foundation models as a function of dataset size used for fine-tuning. For both models used, there is an increase in accuracy with higher number of trajectories used. Further, as expected, the larger pretrained model consistently outperforms the smaller one, albeit by little.
The flattening of the accuracy curve occurring with large amounts of data might be a consequence of the discretization used in fine-tuning, which is limited to a floating number with a sensitivity of 10\textsuperscript{-6}.

\subsection{Generalization}
\begin{figure}[htp]
    \centering
    \includegraphics[trim={0cm 6cm 0cm 7cm},clip,width=0.48\linewidth]{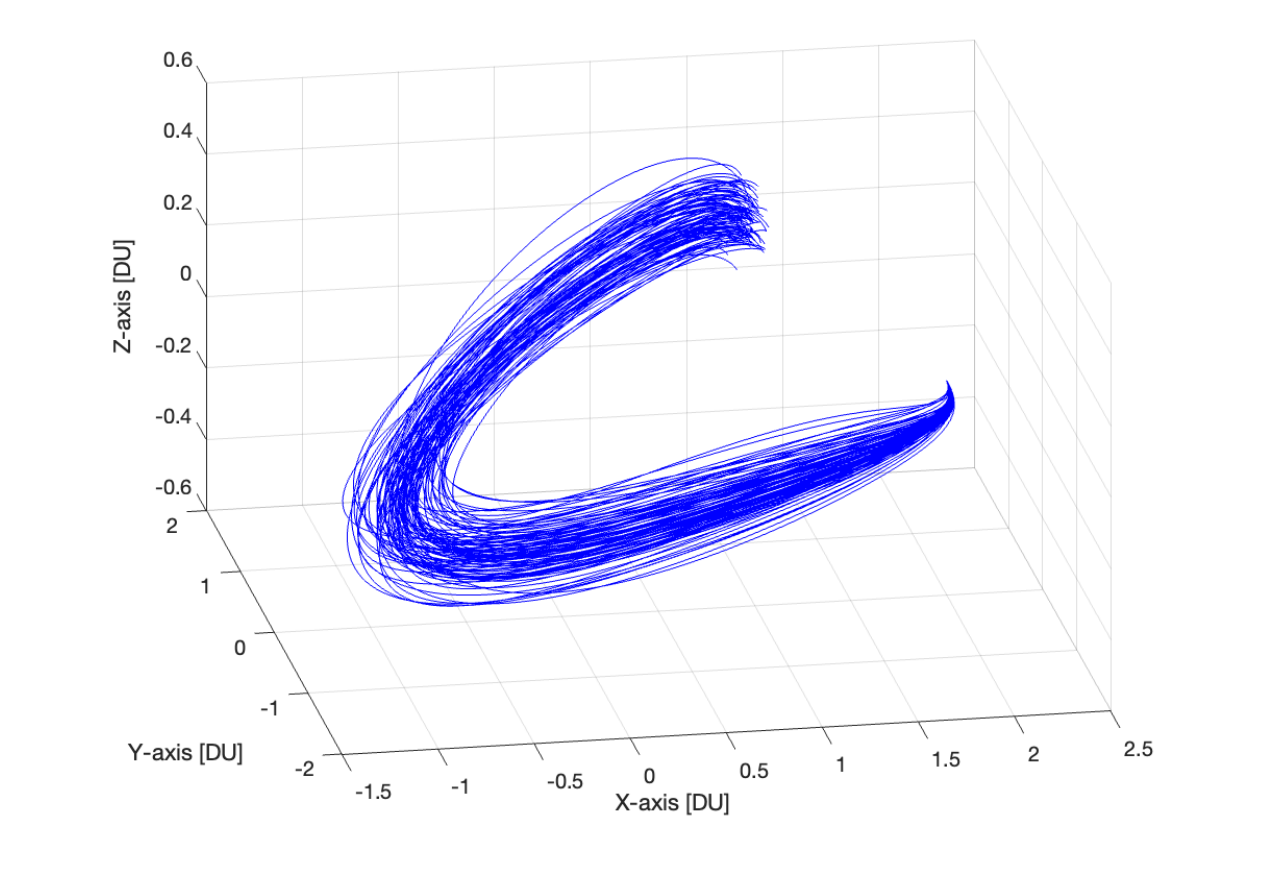}
    \includegraphics[trim={0cm 6cm 0cm 7cm},clip,width=0.48\linewidth]{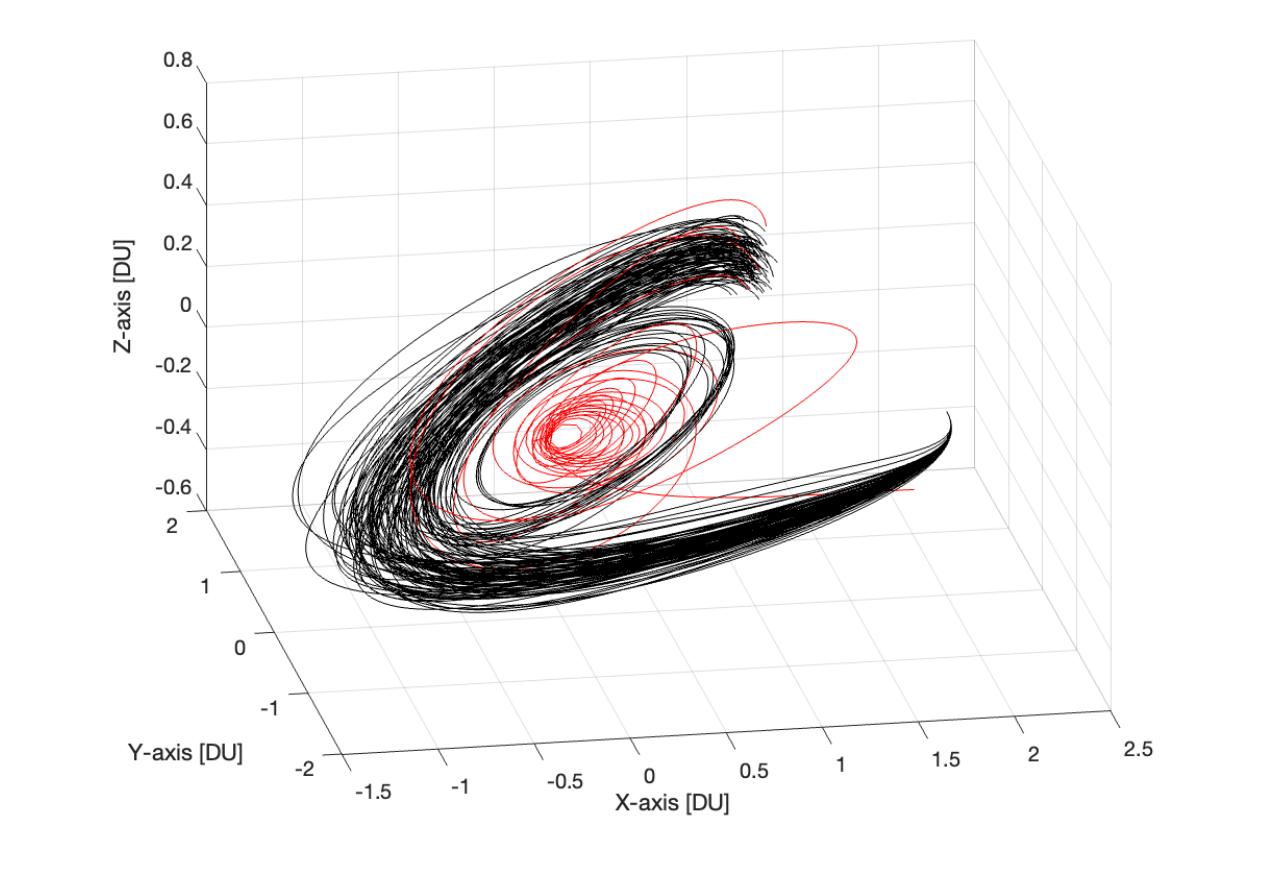}
    \caption{One hundred trajectories with biased initial state distribution. Trajectories guided by the Llama-2-7B-1600 (left), and trajectories computed with the optimizer (right), out of which four do not reach the final state (red), and thirteen more trajectories do not respect the single revolution constraint.}
    \label{fig:orbtransf_1600_7B_biased}
\end{figure}

In the previous section it was shown that for a linear problem the LLMs can generalize well outside of the distribution. While the LLM is not expected to generalize equally as well for this non-convex problem, it is tested outside of the training dataset.
The LLMs' performance of the Llama-2-7B-1600 trajectories has been tested out-of-distribution, by including a bias of half a standard deviation in the initial conditions. Note that, in theory, even when sampling from the same distribution of the dataset, 11.6\% of the trajectories are out-of-distribution, since that is the rate of the non-converging trajectories, which are then not included in the training dataset. Figure~\ref{fig:orbtransf_1600_7B_biased} compares the performance of Llama-2-7B-1600 starting with biased initial conditions with the performance of an optimizer with the same distribution of initial conditions. For the trained LLM, only one trajectory fails to reach the target conditions. On the other hand, the optimizer fails 17 times out of 100. Four trajectories do not converge at all, while 13 more do reach the final target, but without respecting the constraint on the number of revolutions.
When reaching the target, the optimizer achieves the preset tolerance, which is much more accurate than the LLM, which instead, in this case, only reaches an RMSE of 3.876$\times$10\textsuperscript{-3} DU. Hence, accuracy is traded for an increase in robustness with respect to the optimizer.

\begin{figure}[htp]
    \centering
    \includegraphics[trim={4cm 8cm 4cm 9cm},clip,width=0.48\linewidth]{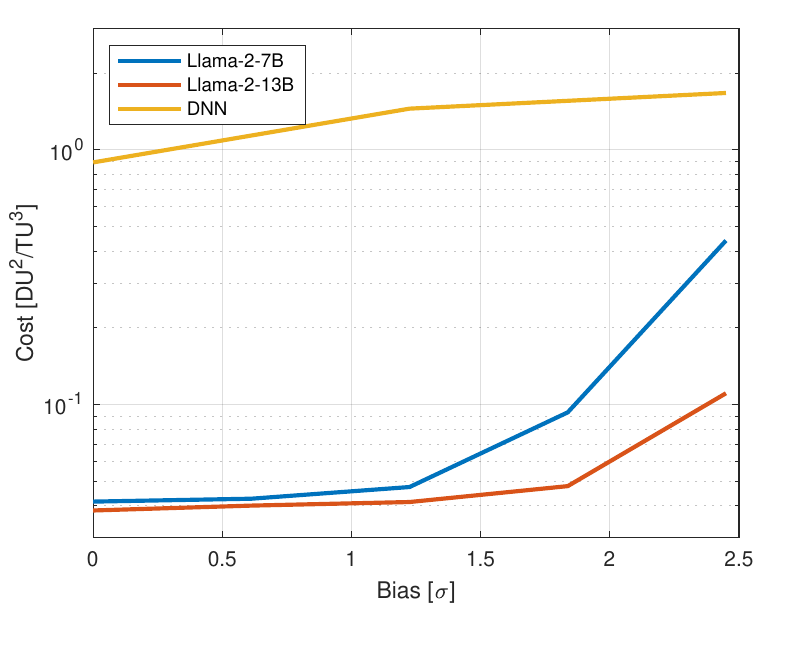}
    \includegraphics[trim={4cm 8cm 4cm 9cm},clip,width=0.48\linewidth]{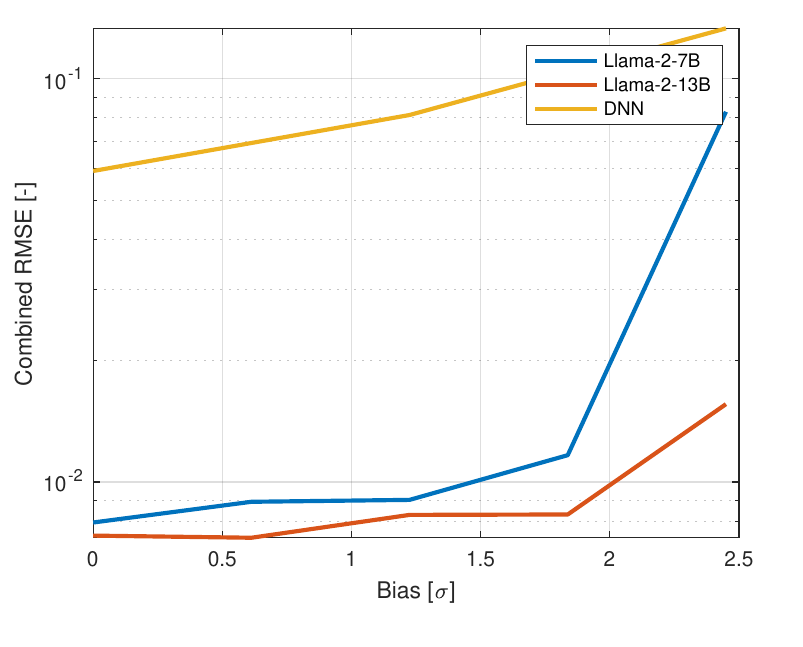}
    \caption{Performance of LLMs compared to a DNN as a function of bias.}
    \label{fig:orbit_generalization_cost}
\end{figure}

The generalization capabilities of LLMs are summarized in Fig.~\ref{fig:orbit_generalization_cost}, which shows the accuracy and cost of Llama-2-7B, Llama-2-13B, and a deep neural network (DNN), all trained with 1,600 trajectories in the dataset, as a function of bias (for the DNN, some outliers not reaching close enough to the target were removed). All results are the statistics from 100 Monte Carlo runs. Both LLMs generalize better than the DNN, and Llama-2-13B generalizes better than Llama-2-7B, as expected. That is especially true for larger bias.

\FloatBarrier
\section{Trajectory Transfer in CR3BP}

With the growing interest in cislunar space, the circular restricted 3-body problem (CR3BP) is considered here as another scenario where LLMs could support autonomous operations in the future. The equations of motion in the CR3BP can be written in a non-dimensional, rotating reference frame where the primary bodies are fixed on the x-axis. The dynamics are governed by the following equations:

\[
    \ddot{x} - 2\dot{y} = \frac{\partial \Omega}{\partial x}, \:
    \ddot{y} + 2\dot{x} = \frac{\partial \Omega}{\partial y}, \:
    \ddot{z} = \frac{\partial \Omega}{\partial z},
\]
where the effective potential \(\Omega\) is given by
\[
\Omega = \frac{1}{2}(x^2 + y^2) + \frac{1 - \mu}{r_1} + \frac{\mu}{r_2},
\]
with \(\mu\) being the mass ratio of the Moon and the Earth, and \(r_1\) and \(r_2\) are the distances from the spacecraft to the primary bodies.

Trajectory planning and optimization in the cislunar domain usually focus on the area close to Lagrange points. The dynamics around the Lagrange points can be further simplified as linear dynamics. Specifically, for the L1 point, the state-space representation of the linearized dynamics can be given by the matrices \(A\) and \(B\).

\[
A = \begin{bmatrix}
    0 & 0 & 0 & 1 & 0 & 0 \\
    0 & 0 & 0 & 0 & 1 & 0 \\
    0 & 0 & 0 & 0 & 0 & 1 \\
    1 + 2\alpha & 0 & 0 & 0 & 2 & 0 \\
    0 & 1 - \alpha & 0 & -2 & 0 & 0 \\
    0 & 0 & -\alpha & 0 & 0 & 0
\end{bmatrix},
\:
B = \begin{bmatrix}
    0 & 0 & 0 \\
    0 & 0 & 0 \\
    0 & 0 & 0 \\
    1 & 0 & 0 \\
    0 & 1 & 0 \\
    0 & 0 & 1
\end{bmatrix},
\]
where the parameter \(\alpha\) is given by the following equation,
\[
\alpha = \frac{1 - \mu_u}{|L_1 + \mu_u|^3} + \frac{\mu_u}{|L_1 - (1 - \mu_u)|^3},
\]
the mass ratio of the Moon and the Earth is \(\mu_u = 0.012150585609624\), and non-dimensional distance to the \(L_1\) point is \(L_1 = 0.8362924\).

The goal of the transfer trajectory is to reach Lagrange point L1 under a given control constraint. This means the boundary conditions are specified by the given initial conditions $\mathbf{x}_{\text{initial}}$ and final conditions $\mathbf{x}_{\text{final}}$, and the controls are bound by some maximum and minimum values. Under this linearized dynamics, the desired target state $\mathbf{x}_{\text{final}}$ is the L1 position at the origin $[0,0,0]^\top$. Furthermore, we choose maximum and minimum values of $1\times10^{-4}$ and $-1\times10^{-4}$, respectively, for each control component.

We formulate the following general linear programming problem with $N$ discretization points:
\begin{align*}
    \text{minimize} \quad & J \\
    \text{subject to} \quad & \mathbf{x}_{k+1} = A\mathbf{x}_k + B\mathbf{u}_k, \quad k = 0, \ldots, N-1, \\
    & \mathbf{x}_0 = \mathbf{x}_{\text{initial}}, \\
    & \mathbf{x}_N = \mathbf{x}_{\text{final}}, \\
    & |\mathbf{u}_{k,d}| \leq 1 \times 10^{-4}, \quad k = 0, \ldots, N-1;\, d = x, y, z.
\end{align*}

As we are primarily interested in feasible trajectories, we do not consider an objective function $J$ for this scenario. Feasible trajectories and the corresponding controls are generated from the linear program and collected as a database for finetuning LLMs. Since LLMs are generally good at sequence prediction \cite{mirchandani2023large}, we employed a simple conversion for the control to comply with sequence prediction.
\[
u_{\text{llm}} = 
\begin{cases} 
1 & \text{if } u < 0, \\
2 & \text{if } u \geq 0.
\end{cases}
\]
Hence, in this case, the output is binary.
The states and their corresponding control information are then collected and compiled into training data for LLMs. We separate the training data into 10 trajectories, 100 trajectories, 500 trajectories, and 1000 trajectories.

\begin{figure}
    \centering
    \includegraphics[width=0.95\linewidth]{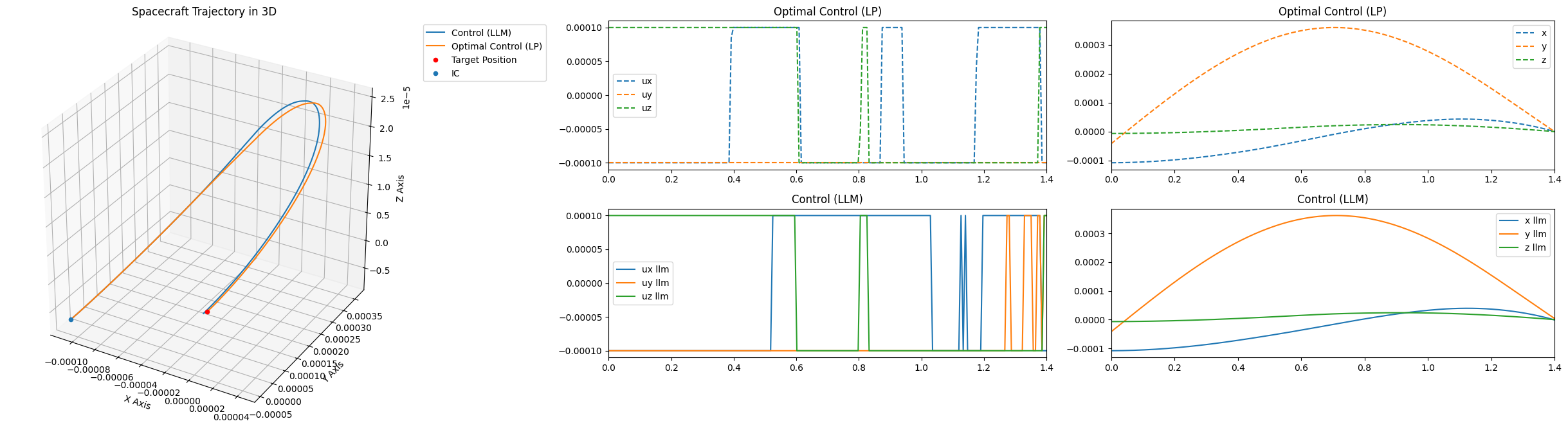}
    \caption{Single trajectory transfer from linear programming and finetuned Llama3-8b.}
    \label{fig:single_traj_llama}
\end{figure}

By fine-tuning a single LLM using existing trajectories, the LLM can learn the correct way of determining the transfer trajectory, as shown in Fig.~\ref{fig:single_traj_llama}. The LLM could guide a spacecraft to L1 in a similar pattern as solving the corresponding linear program. Besides a similar isolated evolution in each direction shown in the right of Fig.~\ref{fig:single_traj_llama}, we can also observe a similar pattern in the control signal in the middle of Fig.~\ref{fig:single_traj_llama}. There is a difference in the control signal, which we expect to see, as the LLM does not directly follow the linear programming result but generates its own control based on the state inputs.

\begin{figure}[htp]
    \centering
    \includegraphics[width=0.7\linewidth]{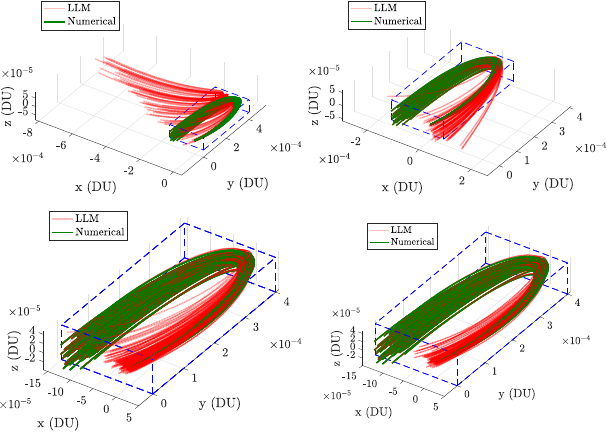}
    \caption{Finetuned Llama2-7B performance on trajectory optimization. The numbers of trajectories for finetuning are, from the top left in a clockwise direction, 10, 100, 500, and 1000. The green trajectories are the numerically solved trajectories, and the red trajectories are from the finetuned Llama2-7b. A cuboid (marked by blue edges) of the same size is used as a reference to compare across different cases.}
    \label{fig:llama2-7b}
\end{figure}

\begin{figure}[htp]
    \centering
    \includegraphics[width=0.7\linewidth]{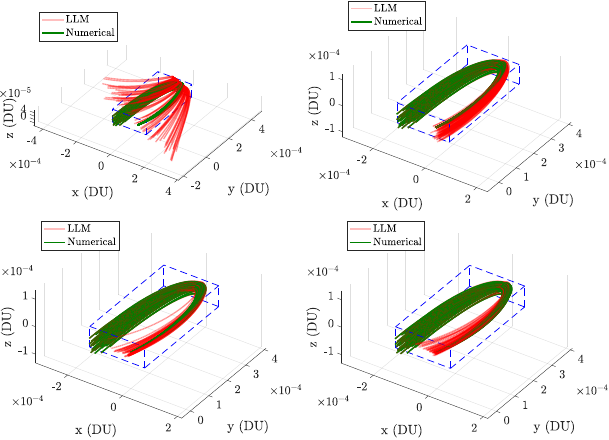}
    \caption{Finetuned Llama2-13B performance on trajectory optimization. The numbers of trajectories for finetuning are, from the top left in a clockwise direction, 10, 100, 500, and 1000. The green trajectories are the numerically solved trajectories, and the red trajectories are from the finetuned Llama2-13b. A cuboid (marked by blue edges) of the same size is used as a reference to compare across different cases.}
    \label{fig:llama2-13b}
\end{figure}

\begin{figure}[htp]
    \centering
    \includegraphics[width=0.7\linewidth]{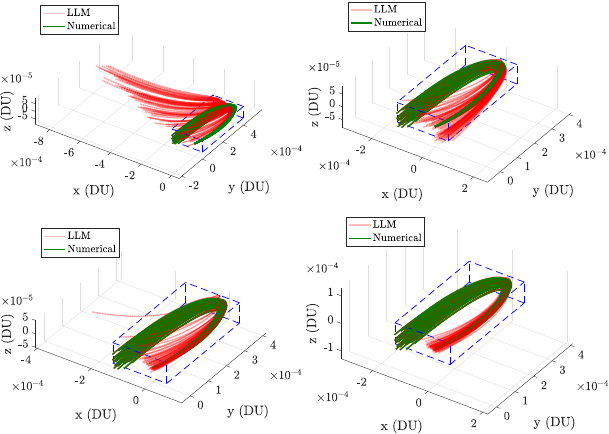}
    \caption{Finetuned Llama3-8b performance on trajectory optimization. The numbers of trajectories for finetuning are, from the top left in a clockwise direction, 10, 100, 500, and 1000. The green trajectories are the numerically solved trajectories, and the red trajectories are from the finetuned Llama3-8b. A cuboid (marked by blue edges) of the same size is used as a reference to compare across different cases.}
    \label{fig:llama3-8b}
\end{figure}

\begin{figure}
    \centering
    \includegraphics[width=0.6\linewidth]{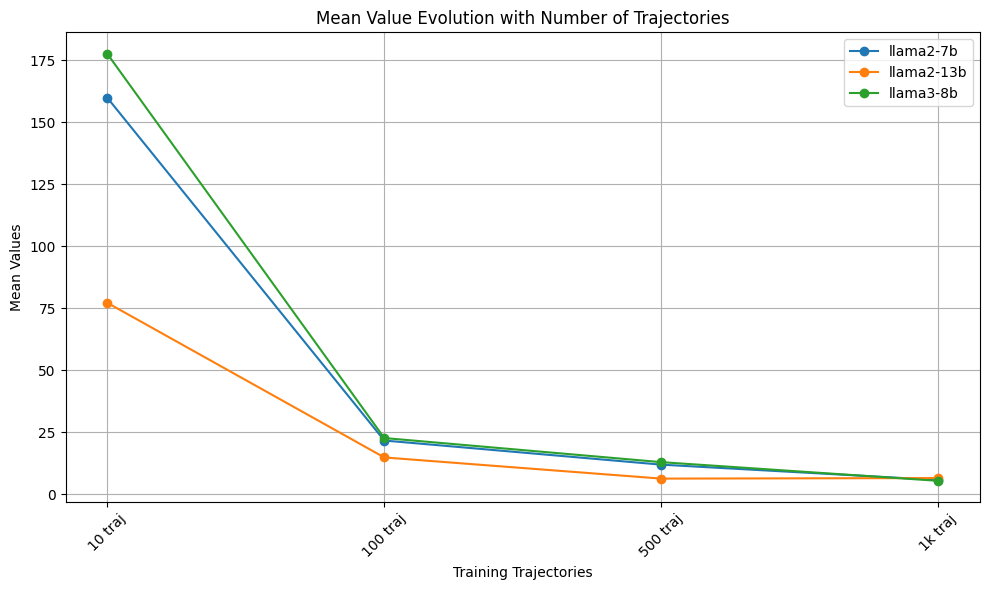}
    \caption{Mean distance values for Llama2-7b, Llama2-13b, and Llama3-8b at different numbers of training trajectories. The accuracy of reaching the final distance increases with more finetuning trajectories. Llama2-13B improves faster with the increase in training samples, while all models reach similar accuracy when 1000 finetuning examples are provided.}
    \label{fig:mean_distance}
\end{figure}

We experiment with three different LLM models using the same collection of trajectory databases. The three LLMs are Llama2-7b, Llama2-13b, and Llama3-8b. There are differences in size as well as generation among the three LLMs we selected. Their performance on 60 testing cases (not included in any of the training databases) is illustrated in Fig.~\ref{fig:llama2-7b}, Fig.~\ref{fig:llama2-13b}, and Fig.~\ref{fig:llama3-8b}. The statistical performance is listed in Tab.~\ref{tab:performance_comparisoncr3bp}, and their mean distance performance is shown in Fig.~\ref{fig:mean_distance}. We observe a clear trend of increasing performance as we increase the size of the finetuning database. Additionally, the LLM with a larger size, Llama2-17b in Fig.~\ref{fig:mean_distance}, has a faster learning performance compared to the smaller LLMs when there are fewer finetuning examples. This indicates higher sampling efficiency, which is generally regarded as beneficial for learning. However, as the size of the database increases, all LLMs converge to a similar performance.

\begin{table}[h]
    \centering
    \begin{tabular}{|c|c|c|c|c|c|c|c|c|c|c|c|}
        \hline
        \textbf{Data size} & \multicolumn{3}{c|}{\textbf{llama2-7b}} & \multicolumn{3}{c|}{\textbf{llama2-13b}} & \multicolumn{3}{c|}{\textbf{llama3-8b}} \\
        \hline
        unit km & \textbf{Min} & \textbf{Mean} & \textbf{Max} & \textbf{Min} & \textbf{Mean} & \textbf{Max} & \textbf{Min} & \textbf{Mean} & \textbf{Max} \\
        \hline
        10 traj & 2.18 & 159.86 & 224.56 & 12.38 & 77.19 & 156.54 & 9.90 & 177.57 & 364.60 \\
        \hline
        100 traj & 0.67 & 21.67 & 72.59 & 0.35 & 14.85 & 29.30 & 0.91 & 22.67 & 62.15 \\
        \hline
        500 traj & 0.66 & 11.94 & 60.58 & 0.71 & 6.32 & 37.31 & 0.52 & 12.95 & 130.12 \\
        \hline
        1000 traj & 0.46 & 5.75 & 21.74 & 0.50 & 6.54 & 29.30 & 0.39 & 5.42 & 25.09 \\
        \hline
    \end{tabular}
    \caption{Performance comparison of LLMs and finetuning data size. 60 initial conditions out of the training are used to test the accuracy in terms of the distance difference between the final state and the target state. In the table, the triplet (min, mean, max) is used to represent the minimum, mean, and maximum distance.}
    \label{tab:performance_comparisoncr3bp}
\end{table}

\section{3 Degrees of Freedom Powered Descent Guidance}
Lastly, we solve the convex fuel-optimal, 3 degrees-of-freedom (DoF) landing problem, defined using lossless convexification (LCvx). The 3 DoF powered descent guidance problem determines the trajectory, which minimizes the thrust input required to achieve pinpoint landing subject to a set of dynamics, state and control, and boundary constraints.
In this scenario, the gravity is constant, and we are targeting a specific point on a flat surface as in Briden et al.~\cite{briden2024improving}.

This benchmark problem complements the orbital transfer and cislunar applications. LCvx is solved using a direct method that discretizes and solves the parameter optimization problem by numerical optimization, in this case, using an interior point method (IPM) solver. This contrasts the orbital transfer problem, which is indirectly solved by solving a two-point boundary value problem. Benchmarking the LLM policies learned using indirect and direct optimal control problem formulations enables practitioners to gauge the accuracy of LLMs on each problem type.

A full derivation of LCvx is defined in Malyuta et al. \cite{Malyuta2021}. The 3 DoF powered descent guidance second-order cone program (SOCP) is defined in Eqs. (\ref{eq: objective function convex})-(\ref{eq: boundary conditions final convex}):

\begin{subequations}
\begin{equation}
\min_{\xi,u,t_f} \int_{0}^{t_f} \xi(t) \mathrm{d} t
\label{eq: objective function convex}
\end{equation}

\begin{equation}
\textrm{s.t.} \quad \dot{r}(t) = v(t), 
\label{eq: dynamics r convex}
\end{equation}

\begin{equation}
\quad \dot{v}(t) = g + u(t) - \omega^{\times} \omega^{\times} r(t) - 2 \omega^{\times} v(t), 
\label{eq: dynamics v convex}
\end{equation}

\begin{equation}
\quad \dot{z}(t) = - \alpha \xi(t), 
\label{eq: dynamics z convex}
\end{equation}

\begin{equation}
\quad \mu_{\min}(t) [1 - \delta z(t) + \frac{1}{2} \delta z(t)^2] \leq \xi(t), 
\label{eq: control constraints lower convex}
\end{equation}

\begin{equation}
\quad \mu_{\max}(t) [1 - \delta z(t)] \geq \xi(t), 
\label{eq: control constraints upper convex}
\end{equation}

\begin{equation}
\quad ||u(t)||_2 \leq \xi(t), 
\label{eq: control constraints norm convex}
\end{equation}

\begin{equation}
\quad u(t)^T \hat{e}_z \geq \xi \cos(\gamma_p), 
\label{eq: control constraints convex}
\end{equation}

\begin{equation}
\quad H_{\text{gs}} r(t) \leq h_{\text{gs}}, 
\label{eq: state constraints glide convex}
\end{equation}

\begin{equation}
\quad ||v(t)||_2 \leq v_{\max}, 
\label{eq: state constraints velocity convex}
\end{equation}

\begin{equation}
\quad \ln(m_{\text{dry}}) \leq z(t_f), 
\label{eq: state constraints dry mass convex}
\end{equation}

\begin{equation}
\quad z_0(t) \leq z(t) \leq \ln(m_{\text{wet}} - \alpha \rho_{\min} t), 
\label{eq: state constraints z bounds convex}
\end{equation}

\begin{equation}
\quad r(0) = r_0, \; v(0) = v_0, \; z(0) = \ln(m_{\text{wet}}), 
\label{eq: boundary conditions initial convex}
\end{equation}

\begin{equation}
\quad r(t_f) = v(t_f) = 0. 
\label{eq: boundary conditions final convex}
\end{equation}
\end{subequations}

Equations (\ref{eq: objective function convex})-(\ref{eq: boundary conditions final convex}) map to the original non-convex formulation through the process of lossless convexification. Where $\alpha$ is the fuel consumption rate, $\gamma_p$ is the maximum allowable tilt angle, $\omega^\times$ is the skew-symmetric matrix for planetary angular velocity, $\rho$ is the thrust magnitude bound, $\xi$ is the thrust vector control input slack variable divided by the vehicle's mass, $g$ is the planetary gravitational acceleration, $H_{\text{gs}}$ is the glideslope constraint matrix, $h_{\text{gs}}$ is the glideslope constraint vector, $m$ is the spacecraft's mass, $r$ is the spacecraft's position, $t_f$ is the final time, $v$ is the spacecraft's velocity, $z$ is the natural log of the vehicle's mass, $u$ is the thrust vector control input divided by the vehicle's mass, and $\hat{e}_z$ is the local vertical unit vector at the landing sit. A slack variable removes the nonconvex lower bound on thrust. Then, the variables $\xi$, $u$, and $z$ are used to approximate nonlinear functions of mass. A new objective function, Eq. (\ref{eq: objective function convex}), maximizes final mass, equivalent to minimizing fuel consumption. A Taylor series approximation is applied to the thrust bounds to transform the convex exponential cone constraint into a second-order cone. An additional constraint, Eq. (\ref{eq: state constraints z bounds convex}), is then added to ensure the maximum fuel rate is not exceeded. To formulate this problem as a free final time problem, a line search or other search method is often used to find a feasible and fuel-optimal final time \cite{Malyuta2021}.

The fine-tuning dataset corresponds to the test data from Briden et al.\cite{briden2024improving}, where uniform sphere sampling was performed on the initial position and velocity, engine angle, pointing angle, and glideslope angle. Using the optimal control problem's formulation and the set of initial conditions, the fine-tuned LLM will iteratively define the next action that satisfies the fine-tuned model's knowledge of fuel optimal cost subject to both the dynamics and constraints.
Results for the landing problem are shown in the next section, where the performance of an LLM only trained for landing is compared to an LLM trained for both landing and orbit transfer. For this problem, the position and velocity inputs have between 8 and 10 significant digits, and the output thrust has between 10 and 11 significant digits. 

\section{LLMs Fine-Tuned for Multiple Space Problems}

\begin{figure}[htp]
    \centering
    \includegraphics[trim={2cm 6cm 2cm 7cm},clip,width=0.6\linewidth]{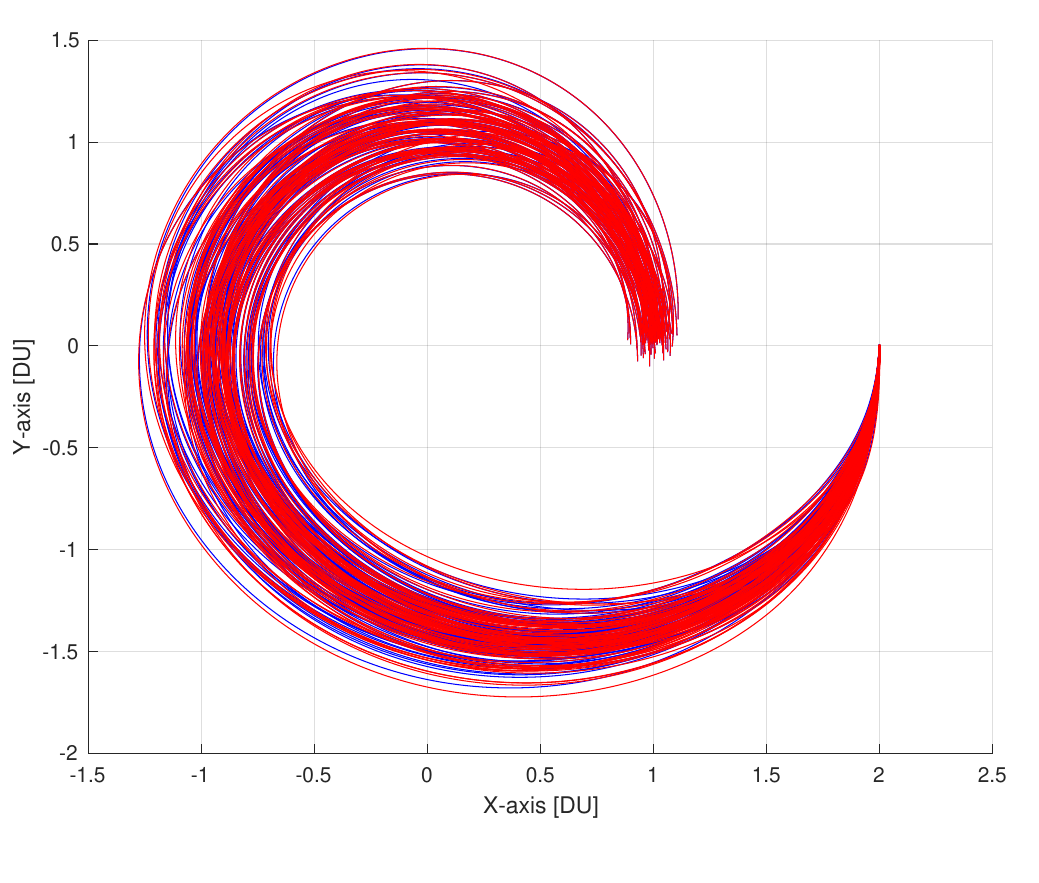}
    \caption{Comparison between OT-LLM (blue) and LOT-LLM (red).}
    \label{fig:multitaskingorbit}
\end{figure}

\begin{figure}[htp]
    \centering
    \includegraphics[trim={4cm 8cm 4cm 9cm},clip,width=0.48\linewidth]{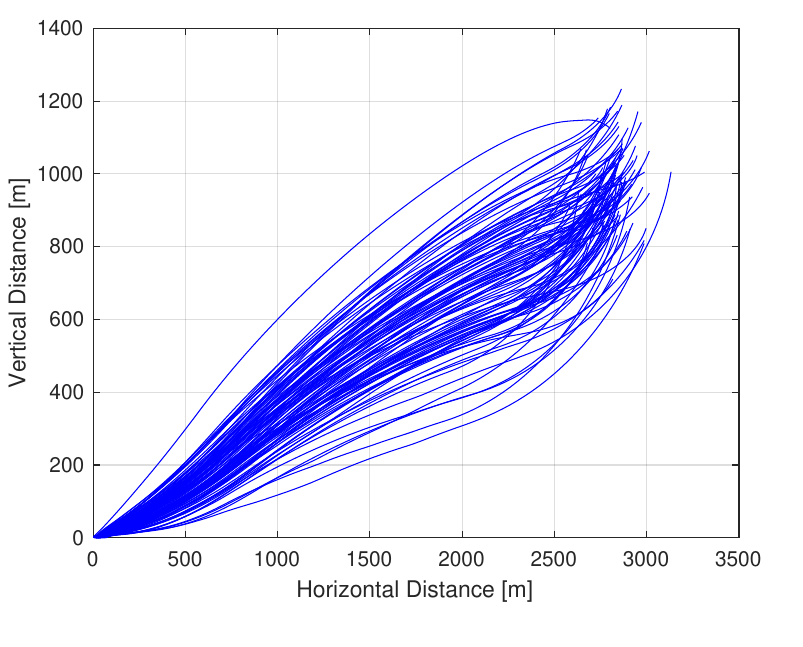}
    \includegraphics[trim={4cm 8cm 4cm 9cm},clip,width=0.48\linewidth]{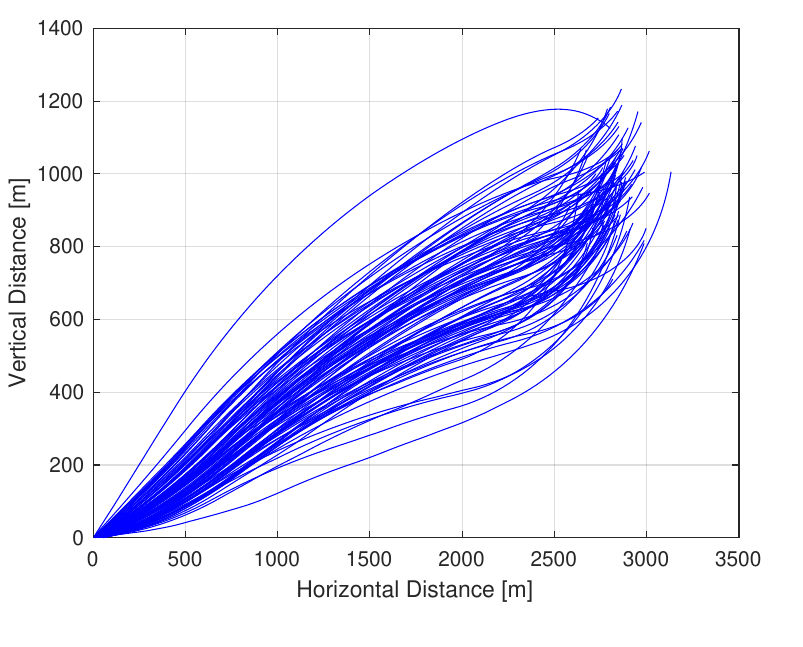}
    \caption{Landing distance during the powered descent phase: L-LLM on the left, and LOT-LLM on the right.}
    \label{fig:landing_distance}
\end{figure}

\begin{figure}[htp]
    \centering
    \includegraphics[trim={3cm 8cm 3cm 9cm},clip,width=0.7\linewidth]{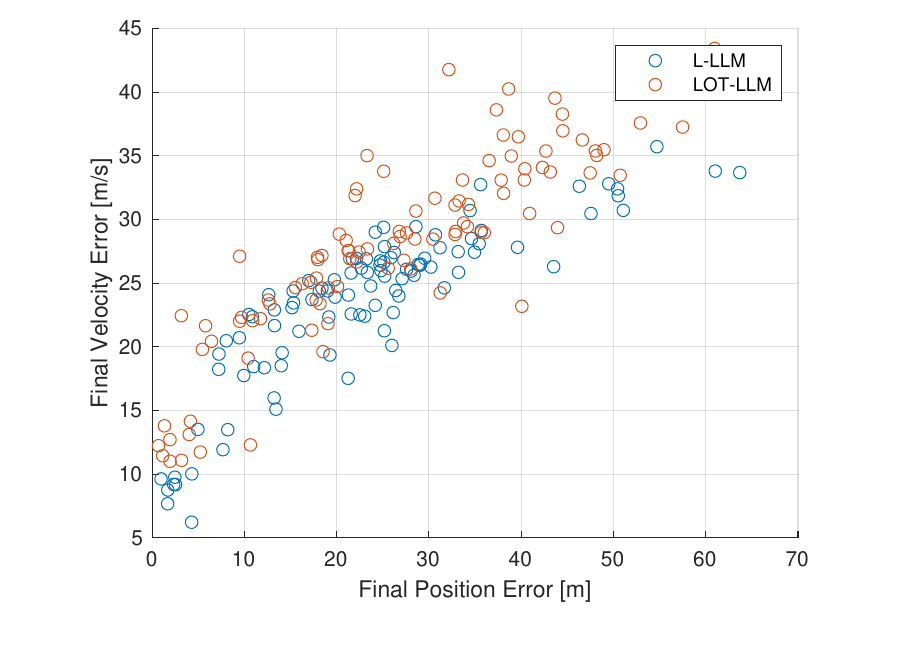}
    \caption{Landing position and velocity error.}
    \label{fig:landing_final}
\end{figure}

\begin{figure}[htp]
    \centering
    \includegraphics[trim={3.5cm 8cm 4cm 9cm},clip,width=0.49\linewidth]{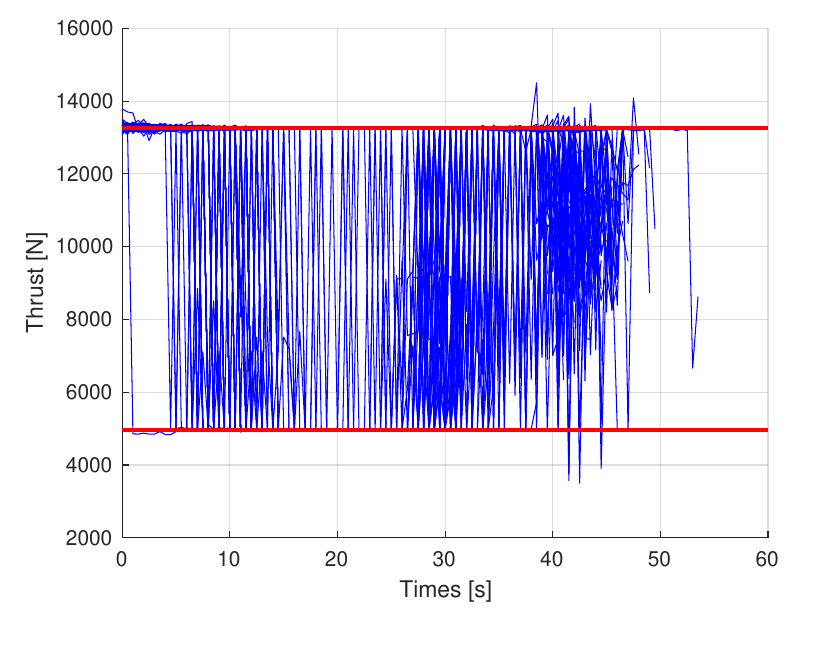}
    \includegraphics[trim={3.5cm 8cm 4cm 9cm},clip,width=0.49\linewidth]{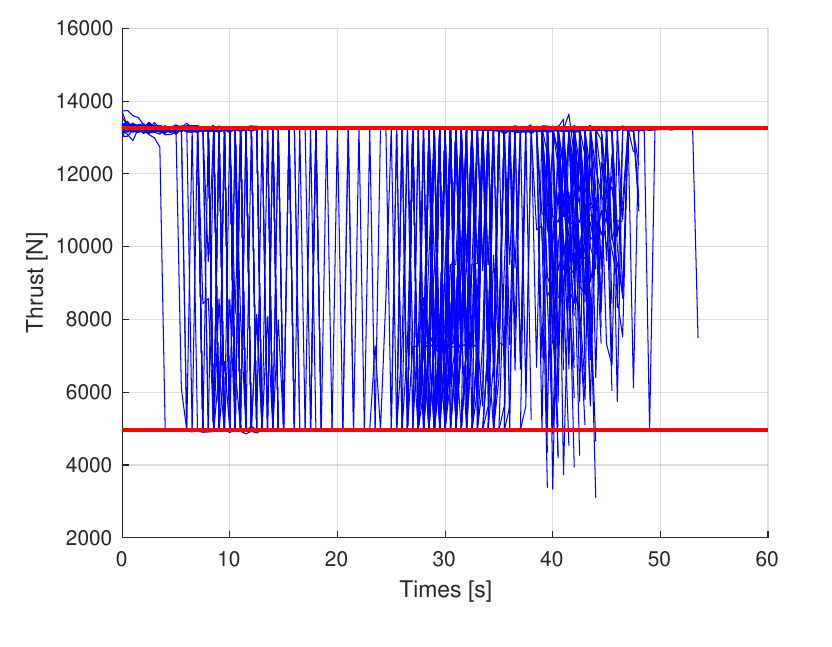}
    \caption{Thrust during the powered descent phase: L-LLM on the left, and LOT-LLM on the right.}
    \label{fig:landing_thrust}
\end{figure}

Two LLMs have been trained to solve the landing problem described in the previous section: both are Llama-2-7B, but one has  been fine-tuned to solve the orbital transfer problem as well, and has thus been fine-tuned to solve the landing and orbit transfer (LOT-LLM), whereas the other one has only been fine-tuned for landing (L-LLM). On the orbit transfer, the LLM-LOT performs very similarly to the equivalent LLM trained only for orbit transfer (OT-LLM): the RMSE in final conditions, and the RMS in cost, only differ by 2\% between the two, one performing better in terms of cost and the other performing better in terms of accuracy. Those results are shown in Fig.~\ref{fig:multitaskingorbit}.
The goal of this section is, in addition to evaluating the performance of the LLM for the landing problem, to define the ability of the same fine-tuned LLM to learn to solve multiple problems.
Figure~\ref{fig:landing_distance} shows the horizontal and vertical distance during the trajectory. The results from the two LLMs look qualitatively rather similar. 
Figure~\ref{fig:landing_final} shows in more detail the final position and velocity errors. In this case, there is some difference between the results of the two LLMs. Specifically, the best performing cases with LLM fine-tuned only on the landing problem are more accurate than the samples obtained with the LLM that had been previously fine-tuned to solve the orbit transfer too. However, on average, the performance is similar.
Finally, Fig.~\ref{fig:landing_thrust} compares how much the thrust constraint is respected by the two LLMs. Note that, while the problem is bang-bang in continuous-time, since the problem is discretized the optimal thrust is not necessarily always the minimum or the maximum allowed; however, that is generally the case for a majority of time-steps.
In both cases, the thrust limit constraint is respected for a majority of guidance calls; however, there is a deterioration towards the end, likely due to the fact that data is limited for perturbations occurring at the end of the trajectory.
Overall, the LLMs trained for landing result in relatively larger errors than the LLMs trained for orbital transfer. This is likely due to the bang-bang nature of the problem, which usually results in more unstable trajectories.

\section{Conclusions}
This work shows the viability of exploiting fine-tuned language models for space systems controls. LLMs are data efficient also during fine-tuning, even if the task they are trained for is likely extremely different from anything they have been pre-trained for. This is likely due to the fact that FMs can be seen as general pattern recognition machines\cite{mirchandani2023large}. While most control works involving language agents produce outputs that are either binary or single-digit, this paper shows that FMs can produce outputs that may be multi-dimensional vector, each entry containing up to 10 significant digits. For a wide range of problems, including linear unconstrained, linear constrained, convex and constrained, and nonlinear unconstrained, the models better performing in language tasks also learn faster. This further indicates the possibility that the pattern recognition ability of FMs is connected with their language performance, and that pre-training aids even for tasks unrelated from the pre-training data itself. While not being as accurate, in the nonlinear cases the LLM guidance is more robust than even the optimizer used to generate the original dataset used for fine-tuning. That is not the case for the convex problems instead, since in such cases convergence of the optimizer is guaranteed. Additionally, the optimal solution for the evaluated convex problem is bang-bang, in which perturbations, and the inaccuracies introduced by the FMs, can destabilize the system. It is also shown that LLMs generalize well outside of the fine-tuning data, and can perform better than traditional AI when tested outside of the training distribution. Finally, the same LLM can be fine-tuned with data from different problems, so that the same system can later be used to control the spacecraft during different phases of a mission. We thus expect that future LLMs will be even more sample efficient. This paper is a first step towards the development of a general spacecraft controller.

\section{Acknowledgments}
Research was sponsored by the Department of the Air Force Artificial Intelligence Accelerator and was accomplished under Cooperative Agreement Number FA8750-19-2-1000. The views and conclusions contained in this document are those of the authors and should not be interpreted as representing the official policies, either expressed or implied, of the Department of the Air Force or the U.S. Government. The U.S. Government is authorized to reproduce and distribute reprints for Government purposes notwithstanding any copyright notation herein. 
This work was supported in part by a NASA Space Technology Graduate Research Opportunity 80NSSC21K1301 and the National Science Foundation under award NSF-PHY-2028125.



\end{document}